\documentclass[runningheads]{llncs}
\usepackage{graphicx}
\usepackage{color, colortbl}
\definecolor{LightGray}{gray}{0.9}
\usepackage{dsfont}
\usepackage{graphicx}
\usepackage{amsmath}
\usepackage{amssymb}
\usepackage{booktabs}
\usepackage{multirow}
\usepackage{paralist}
\usepackage{array}
\usepackage[normalem]{ulem}
\usepackage[accsupp]{axessibility}
\usepackage{soul}
\usepackage{url}
\usepackage{color, colortbl}
\usepackage{relsize}
\usepackage{subcaption}
\usepackage{xspace}
\usepackage{caption}
\usepackage[pagebackref,breaklinks,colorlinks,bookmarks=false]{hyperref}

\usepackage[capitalize]{cleveref}
\crefname{section}{Sec.}{Secs.}
\Crefname{section}{Section}{Sections}
\Crefname{table}{Table}{Tables}
\crefname{table}{Tab.}{Tabs.}

\usepackage[usenames,dvipsnames]{xcolor}
\usepackage{suffix}
\newcommand{\authorcomment}[3]{\noindent\textsf{\textcolor{#1}{[\textbf{#2:} \textit{#3}]}}}
\WithSuffix\newcommand\authorcomment*[3]{\textsf{\textcolor{#1}{\textit{#3}}}}

\newcommand{\blockcomment}[1]{}

\newcommand\parahead[1]{\noindent\textbf{#1: }}

\newcommand{\FIXME}[1]{\textcolor{red}{#1}\xspace}

\newcommand{\CITEME}[1]{\FIXME{\cite{CITEME}}\xspace}
\newcommand{\OURS}{ELVIS\xspace}
\newcommand{\ours}{\OURS}
\newcommand{\score}[2]{{#1} \scriptsize{$\pm$#2}}

\newcommand{\pretrain}[0]{pre-train}
\newcommand{\Pretrain}[0]{Pre-train}

\newcommand{\by}{\mathbf{y}}
\newcommand{\bz}{\mathbf{z}}
\newcommand{\bp}{\mathbf{p}}
\newcommand{\bq}{\mathbf{q}}
\newcommand{\bs}{\mathbf{s}}
\newcommand{\cM}{\mathcal{M}}
\newcommand{\cI}{\mathcal{I}}
\newcommand{\cT}{\mathcal{T}}

\makeatother

\begin{document}
\title{ELVIS: Empowering Locality of Vision Language \\Pre-training with Intra-modal Similarity}
\titlerunning{\OURS}
\author{
Sumin Seo\inst{1} \and
JaeWoong Shin\inst{2} \and
Jaewoo Kang\inst{3,4} \and
Tae Soo Kim\inst{2}\and
Thijs Kooi\inst{2}
}
\authorrunning{Seo et al.}
\institute{
Galux Inc. \and
Lunit Inc. \and
Korea University\and
AIGEN Sciences Inc.\\ \email{sm.seo@galux.co.kr, \{jwoong.shin, taesoo.kim, tkooi\}@lunit.io, kangj@korea.ac.kr}
}
\maketitle              %
\begin{abstract}
Deep learning has shown great potential in assisting radiologists in reading chest X-ray (CXR) images, but its need for expensive annotations for improving performance prevents widespread clinical application.
Visual language \pretrain ing (VLP) can alleviate the burden and cost of annotation by leveraging routinely generated reports for radiographs, which exist in large quantities as well as in paired form (image-text pairs).
Additionally, extensions to localization-aware VLPs are being proposed to address the needs for accurate localization of abnormalities for computer-aided diagnosis (CAD) in CXR.
However, we find that the formulation proposed by locality-aware VLP literature actually leads to a loss in spatial relationships required for downstream localization tasks. 
Therefore, we propose Empowering Locality of VLP with Intra-modal Similarity, \ours, a VLP aware of intra-modal locality, to better preserve the locality within radiographs or reports, which enhances the ability to comprehend location references in text reports. %
Our locality-aware VLP method significantly outperforms state-of-the art baselines in multiple segmentation tasks and the MS-CXR phrase grounding task.
Qualitatively, we show that \ours focuses well on regions of interest described in the report text compared to prior approaches, allowing for enhanced interpretability.

\end{abstract}
\section{Introduction}
\begin{figure}[t]

	\centering
	\hfill
     \begin{subfigure}[b]{0.49\textwidth}
\captionsetup{justification=centering}
         \centering
         \includegraphics[width=\textwidth]{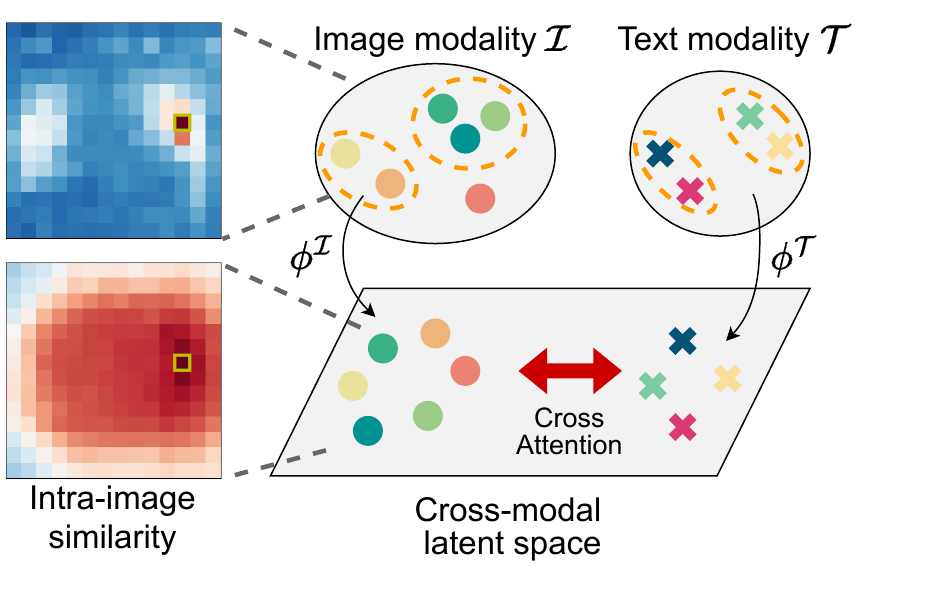}
         \caption{Loss of intra-image \\ correlation after projection~\cite{muller2021joint}}
         \label{fig:lovt}
     \end{subfigure}
	\hfill
     \begin{subfigure}[b]{0.49\textwidth}
\captionsetup{justification=centering}
         \centering
         \includegraphics[width=\textwidth]{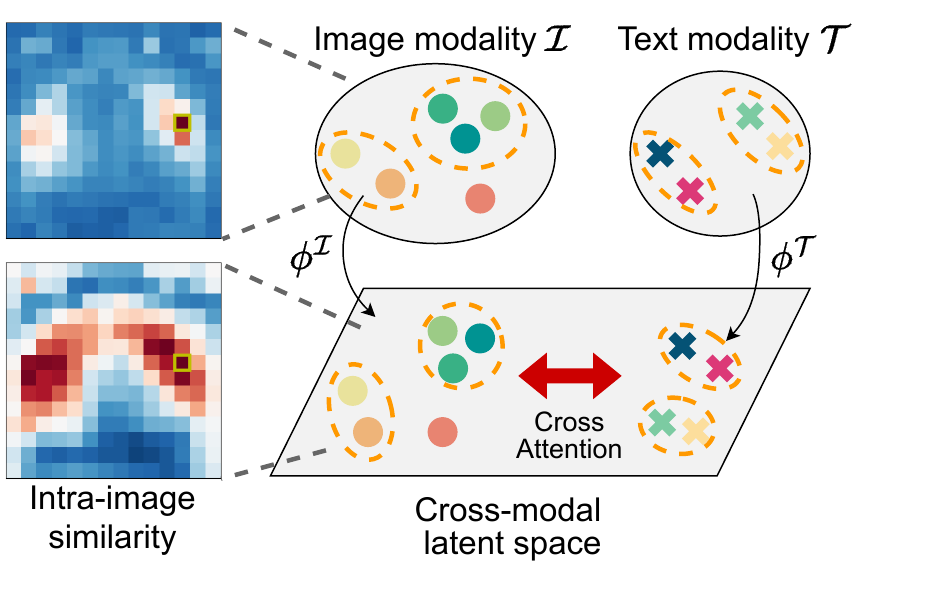}
         \caption{Preservation of intra-image \\ correlation after projection (\OURS)}
         \label{fig:ours}
     \end{subfigure}
	\caption{\textbf{VLP methodologies that project unimodal embeddings to cross-modal space.} Heatmaps are generated from feature maps of chest radiographs. Red color indicates regions similar to the query region highlighted with the yellow box, and blue indicates the less similar regions. In each figure, heatmaps on the top denote the intra-image local similarity map of the yellow box region with the entire image, while the bottom heatmaps describe the possible loss of intra-modal similarity after projection to cross-modal embedding space. %
 } %

\label{fig:intro}
\end{figure}

Deep learning-based computer-aided diagnosis (CAD) systems that assist radiologists in diagnosing chest X-ray (CXR) images require large amounts of annotated data to be trained. 
Unfortunately, collecting data is challenging due to the cost of annotating images with bounding boxes or lesion outlines by board-certified radiologists~\cite{kim2022accurate,willemink2020preparing}. 
Vision language pre-training (VLP) offers an alternative to costly annotations as it enables models to be trained on large amounts of paired image-text data directly. 
Also, VLP models can learn high-level semantic concepts from text descriptions that cannot be learned from images alone and understand the relationship between images and text~\cite{wang-etal-2022-medclip}.
For example, several studies~\cite{jia2021scaling,2020_clip} have demonstrated that VLP models trained on 400 million or even 1 billion noisy image-text pairs collected from the internet perform comparably or even better than supervised baselines on a variety of computer vision benchmarks. Recent studies, such as \cite{wang-etal-2022-medclip,zhang2022contrastive}, have demonstrated the potential of VLP in the CXR domain by using 200k X-ray image-report pairs to improve over commonly used ImageNet initialization~\cite{deng2009imagenet}, on various downstream tasks ranging from classification to zero-shot retrieval.

VLP for CXR must go beyond classification and be extended to detection problems, as CAD in CXR requires accurate localization of abnormalities.
To extend VLP methods to detection problems, recent works~\cite{Huang_2021_ICCV,muller2021joint} introduce contrastive learning on local representations of image patches and text tokens. %
Understanding the inherent spatial redundancy in local representations proves beneficial for learning visual representations for segmentation tasks~\cite{he2022masked}. Furthermore, redefining positive pairs to be situated closer in unified embedding space during contrastive learning has proven advantageous when dealing with loosely correlated pairs~\cite{wu2022data,lee_uniclip}.
While GLoRIA~\cite{Huang_2021_ICCV} primarily focuses on improving visual representations by learning similarities between word and image-attended word features, it does not adequately consider multiple words with similar meanings.
\cite{seibold2022breaking} also suggests multimodal contrastive learning using instance-level image features paired with sentences, and this approach is limited to classification applications. 
LoVT~\cite{muller2021joint} leverages contrastive learning by defining intra-modal pairs in co-embedding space shared by image patches and text tokens. 
While LoVT introduces a distance-based contrastive objective for images, which defines the positive correlation between neighboring regions, it is limited in capturing semantic similarity between distant regions, such as bilateral lung regions. Still, the positive correlation between multiple sentences is constrained when encoding sentence features for this objective.

However, we find that such contrastive formulation actually leads to a loss in the ability to encode spatial relationships required for downstream localization tasks. 
As illustrated in \autoref{fig:lovt}, local representations within the lung are encoded similarly (top row), but the desired similarity disappears after the projection to joint embedding space (bottom row). 
We observe that the distance-based contrastive objective is actually misaligned with how the abnormalities are distributed in a CXR image and leads to a poor embedding for preserving local information. For example, semantically similar image patches may appear on both sides of the lung despite their large spatial distance.

In this paper, we present a novel VLP technique to better preserve intra-modal locality between local representations of CXR and text reports.
Based on our hypothesis, we propose \OURS, Empowering Locality of VLP with Intra-modal Similarity, which models intra-modal similarity as a local contrast objective.
We demonstrate that this formulation allows the model to preserve local embedding distribution within modality and in the co-embedding space, as illustrated in \autoref{fig:ours}.

We \pretrain~our proposed model with the MIMIC-CXR~\cite{2019_mimic_cxr} dataset and evaluate on various downstream localization tasks using four datasets: RSNA Pneumonia~\cite{rsna_pneumonia}, SIIM Pneumothorax~\cite{pneumothorax_segm}, COVID Rural~\cite{tang2020segmentation}, and MS-CXR~\cite{boecking2022making}.
Our model outperforms state-of-the-art VLP baselines on various segmentation tasks and also shows improved comprehension to vision-language correlation over baselines.

We summarize our contributions as follows:
\begin{enumerate}
    \item We propose \ours, a novel VLP framework that adopts two views from semantic units of radiological image-text data. We introduce a local contrastive objective to preserve intra-modal similarity and a global contrastive objective for learning a holistic view.
    \item We demonstrate that the proposed framework outperforms in a variety of downstream applications, which require enhanced comprehension of localized information in x-ray images and link to corresponding textual descriptions.
    \item We provide qualitative analysis which shows that the proposed method learns semantic correlation between image and text by preserving intra-modal locality during pre-training, retaining important contextual and spatial information.
\end{enumerate}

\section{Methods}
To address intra-modal locality in the CXR domain, we propose a novel VLP framework that learns from contrastive learning on both global and local level representations.
We first define global and local representations to contrast in \autoref{sec:arch}, and propose objectives for each contrastive learning in \autoref{sec:training}.

\subsection{Overall architecture}
\label{sec:arch}
\begin{figure}[t]
\centering
    \includegraphics[width=\linewidth]{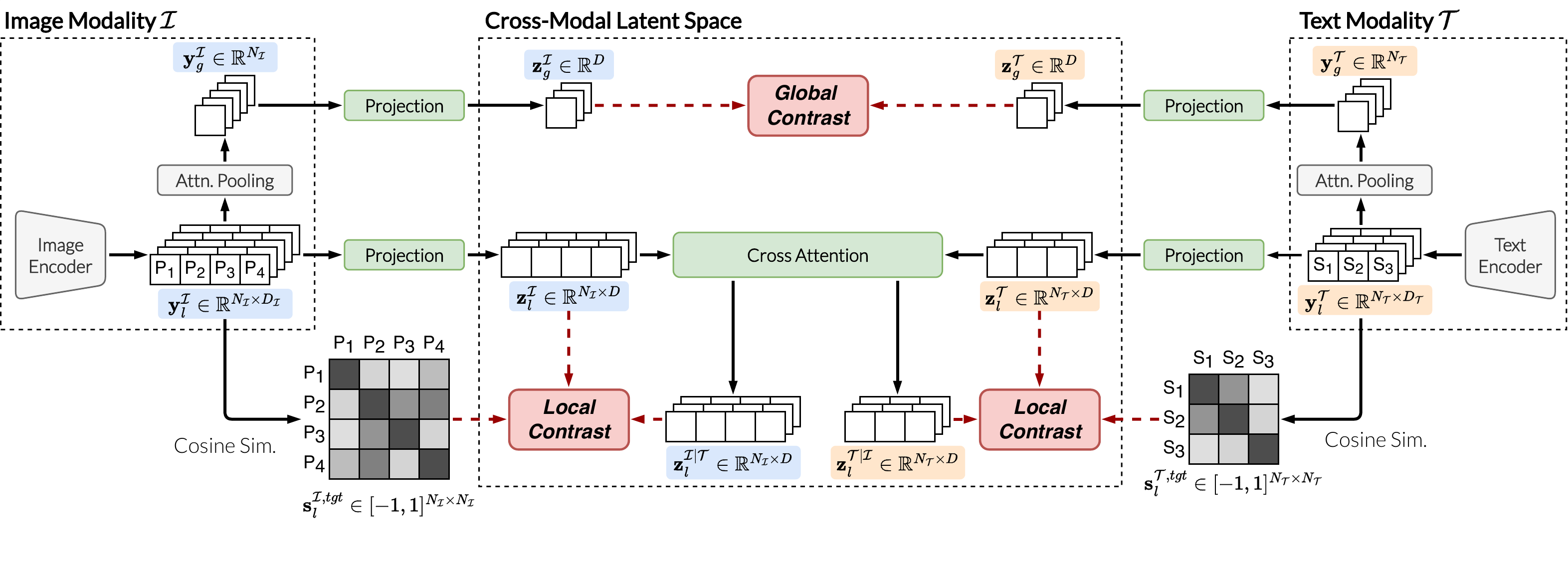}
    \caption{
        \textbf{Overall architecture.} 
        We first encode the image or text input to a local representation ($\mathbf{y}_l^\cM$) and obtain global representation ($\mathbf{y}_g^\cM$) through attention pooling within each modality $\cM$.
        Then, we project embeddings into shared cross-modal latent space by $\mathbf{z}_l^\cM$ and $\mathbf{z}_g^\cM$.
        For global contrastive learning, we contrast $\mathbf{z}_g^\mathcal{I}$ and $\mathbf{z}_g^\mathcal{T}$.
        For local contrastive learning, we first compute cross-attended local representation $\mathbf{z}_l^{\cM|\cM'}$ through cross-attention module using $\mathbf{z}_l^{\cM}$ as query and $\mathbf{z}_l^{\cM'}$ as key and value, and then contrast $\mathbf{z}_l^{\cM}$ and $\mathbf{z}_l^{\cM|\cM'}$.
        }
    \label{fig:arch}
\end{figure}

\parahead{Modality-specific Embeddings} 
We first encode the image or text input into a local representation that summarizes the local semantics, within each modality.
In image modality $\cI$, we obtain a feature map through an image encoder, ResNet-50~\cite{he2015_resnet}, and then define each element of this feature map as a local embedding. Specifically, we use the output of the third stage of ResNet-50.
In text modality $\cT$, we define each sentence in a report as a semantic unit and extract a local embedding for each sentence through a text encoder, such as ClinicalBERT~\cite{alsentzer2019publicly}.
The local embeddings ($\by_l^\cI\in\mathds{R}^{N_\cI\times D_\cI}$ and $\by_l^\cT\in\mathds{R}^{N_\cT\times D_\cT}$) obtained for each modality are then pooled using attention~\cite{2020_clip} to generate a global embedding ($\by_g^\cI\in\mathds{R}^{D_\cI}$ and $\by_g^\cT\in\mathds{R}^{D_\cT}$) for the entire input.

\parahead{Cross-modal Embeddings}
To align embeddings across modalities, we project the modality-specific embeddings $\by_l^\cM$ and $\by_g^\cM$ ($\cM\in\{\cI,\cT\}$) to cross-modal embeddings $\bz_l^\cM\in\mathds{R}^{N_\cM\times D}$ and $\bz_g^\cM\in\mathds{R}^{D}$ over a shared latent space of dimension $D$.
At this point, the global representations of each modality ($\bz_g^\cI$ and $\bz_g^\cT$) can be aligned with each other through InfoNCE loss~\cite{2020_clip}. %
Note, however, that the local representations of each modality ($\bz_l^\cI$ and $\bz_l^\cT$) do not have direct and clear supervision to contrast.

To bypass this, we calculate a cross-attended embedding $\bz^{\cM|\cM'}$ for modality $\cM$ that incorporates information from its counterpart modality $\cM'$ through a co-attention module~\cite{vaswani2017attention}. 
Specifically, we first compute the correlation between $\bz^\cM$ and $\bz^{\cM'}$ using cosine distance. The obtained correlation serves as weights applied to a set of values, which are computed by a linear transform of $\bz^{\cM'}$: %
\begin{equation}
    \bz^{\cM|\cM'} = \left(\frac{(\bz^\cM)^\top \cdot \bz^{\cM'}}{\|\bz^\cM\|\cdot\|\bz^{\cM'}\|}\right)\cdot W_v \bz^{\cM'}
\end{equation}
Here, the weight $W_v$ is shared between image-to-text and text-to-image transforms.
Then, we contrast two representations, $\bz_l^\cM$ and $\bz_l^{\cM|\cM'}$.

\subsection{Training}
\label{sec:training}
For ease of formulation, we define the similarity function between two embeddings $a$ and $b$ of the same dimension as follows: $\text{sim}(a,b)=\frac{a^\top b}{\|a\|\|b\|}$.

\parahead{Global Contrastive Learning}
Following the literature~\cite{2020_clip}, we maximize the similarity between the global representations of paired image and report samples while minimizing that of other samples within a batch.
Formally, we compute a cosine similarity between global representations within a batch by $\bs_{g, ij} = \text{sim}(\bz_{g, i}^\cI,\bz_{g, j}^\cT)$. 
Then, we align global representations through the following contrastive loss,

{
\begin{equation}
    \mathcal{L}_{g,i}^{\cI|\cT}= -\log \frac{\exp \left( \bs_{g,ii}/\tau_g\right)}{\sum_{j=1}^{B}\exp \left(\bs_{g,ij}/\tau_g\right)},\quad
    \mathcal{L}_{g,j}^{\cT|\cI}= -\log \frac{\exp \left( \bs_{g,jj}/\tau_g\right)}{\sum_{i=1}^{B}\exp \left(\bs_{g,ij}/\tau_g\right)}
\end{equation}
}
where $B$ and $\tau_g$ denote the batch size and temperature parameter, respectively. In this work $\tau_g=0.3$ is used.

\parahead{Local Contrastive Learning}
At the local level, we contrast local representations $\bz_l^\cM$ and cross-attended representations $\bz_l^{\cM|\cM'}$ within each image and text sample.
Here, we adopt a contrastive objective supervised by intra-modal similarity based on the idea that local representations that originally have similar meanings would still have similar meanings after cross-attention.
To do so, we first compute two similarities: intra-modal similarity $\bs_l^{\cM,\text{tgt}}=\text{sim}(\by_l^\cM, \by_l^\cM)$ as supervision and similarity between local representations and cross-attended representations by $\bs_l^{\cM,\text{src}}=\text{sim}(\bz_l^\cM, \bz_l^{\cM|\cM'})$. Then we compute probability maps by applying row- or colum-wise softmax with temperature scaling: $\bp_{l}^{\cM,\text{row}}=\text{softmax}^{\text{row}}(\bs_l^{\cM,\text{tgt}}/\tau_l^{\text{tgt}})$, $\bq_l^{\cM,\text{row}}=\text{softmax}^{\text{row}}(\bs_l^{\cM,\text{src}}/\tau_l^{\text{src}})$, and similarly, $\bp_{l}^{\cM,\text{col}}$, $\bq_{l}^{\cM,\text{col}}$. Finally, we align $\bq_{l}^{\cM}$ to $\bp_l^{\cM}$ by cross-entropy loss:
{
\begin{equation}
    \mathcal{L}_l^{\cM}= -\sum_{i=1}^{N_\cM}\sum_{j=1}^{N_\cM} \bp_{l,ij}^{\cM,\text{row}}\log \bq_{l,ij}^{\cM,\text{row}} + \bp_{l,ij}^{\cM,\text{col}}\log \bq_{l,ij}^{\cM,\text{col}}
\end{equation}
}
In this work, $\tau_l^{\text{src}}=0.3, \tau_l^{\text{tgt}}=0.1$ is used.

\parahead{Total Loss}
Finally, the total loss is defined as $\mathcal{L} = \lambda_1\mathcal{L}_{g}^{\cI|\cT} + \lambda_2\mathcal{L}_{g}^{\cT|\cI} + \lambda_3\mathcal{L}_{l}^{\cI} + \lambda_4\mathcal{L}_{l}^{\cT}$ and $[\lambda_1, \lambda_2, \lambda_3, \lambda_4]=[0.25,0.75,0.375,0.375]$ are used in this work.

\parahead{Training Details}
We \pretrain~\ours using a subset of MIMIC-CXR-JPG~\cite{2019_mimic_cxr}, which consists of 377,110 chest X-ray images of 227,827 patients, along with 227,827 associated reports. We filter out lateral and other views to match downstream tasks that only require frontal-view images, resulting in 203,992 frontal-view image-text pairs for training.
We \pretrain~the model for 50 epochs using the Adam~\cite{adam} optimizer with a learning rate of 1e-4 and cosine scheduling~\cite{loshchilov2016sgdr}.

\section{Experiments}
We first evaluate the proposed model using three lesion localization datasets to assess its transferability and localization performance.
Next, we conduct phrase grounding experiments to evaluate how well our model comprehends the relationship between image and text.

\subsection{Baselines}
We compare \OURS with seven baselines: random initialization, ImageNet~\cite{deng2009imagenet} \pretrain ed initialization, five VLP methods (CLIP~\cite{2020_clip}, ConVIRT~\cite{zhang2022contrastive}, LoVT~\cite{muller2021joint}, GLoRIA~\cite{Huang_2021_ICCV}, and BioViL~\cite{boecking2022making}).
ImageNet is a strong baseline in the general domain but is not specifically trained on CXR data. 
In the case of CLIP, we train it on CXR data by ourselves.
ConVIRT and BioVIL are trained on CXR data using global contrastive learning, similar to CLIP, but BioVIL uses a better language understanding model that takes into account the nature of CXR reports. Note that BioVIL addresses both local image features and global pooled features for contrastive learning. We utilize ResNet50~\cite{he2015_resnet} image encoder for all baselines.
GLoRIA and LoVT, on the other hand, are trained using local contrastive learning in addition to global contrastive learning, with each assuming a unit of text local semantics as word-piece and sentence, respectively. 
However, unlike \OURS, these methods do not address intra-modal local similarity.

\subsection{Lesion Segmentation}
We first assess the transferability and localization performance of \OURS and the baselines by transferring to downstream localization tasks.

\parahead{Datasets and evaluation details}
We employ three localization datasets with three different chest abnormalities.
1) The \textbf{RSNA Pneumonia Detection} dataset~\cite{rsna_pneumonia} consists of 26,684 frontal-view chest radiographs and bounding boxes, which indicate the location of pneumonia. We utilize this dataset for a segmentation task using the bounding box annotations as ground truth segmentation targets. To achieve segmentation, we perform linear probing by freezing the \pretrain ed encoder and training only a linear layer after the encoder.
2) The \textbf{COVID Rural Segmentation} dataset~\cite{tang2020segmentation} contains 221 frontal-view chest radiographs with segmentation annotations for detecting COVID-19. For this segmentation task, we perform linear probing as well as fine-tuning the \pretrain ed encoder followed by a U-Net~\cite{ronneberger2015unet} decoder architecture.
3) Lastly, the \textbf{SIIM Pneumothorax Segmentation} dataset \cite{pneumothorax_segm} comprises 12,047 frontal-view chest radiographs and segmentation masks for localizing pneumothorax. We perform fine-tuning with a U-Net decoder for this segmentation task.

For all datasets, we resize images to $224\times224$ and adopt image augmentations, including scaling and rotation following \cite{Huang_2021_ICCV}.
We train segmentation models for 50 epochs while carefully selecting an appropriate learning rate (in the range of [3e-5, 1e-2]) and scheduling strategy that considers overfitting. 
The detailed hyperparameters for downstream training can be found in the supplementary material.
All downstream experiments are repeated four times, and the mean Dice score with a 95\% confidence interval is reported.

\begin{table}[t]
{
\centering
\caption{\textbf{Dice scores (\%) on four downstream segmentation tasks.} 
The best score on each column is highlighted in bold, and the second best is underlined.
\dag: We use pre-trained weights provided by the authors, whereas the other baselines are reproduced by ourselves. Note that GLoRIA and BioViL are pre-trained with higher image resolution, $299\times299$ and $512\times512$, respectively.
}
\label{tab:main}

    \centering
    \begin{tabular}{lrrrrrrrr}
    \toprule
    \multicolumn{1}{c}{\multirow{2}[2]{*}{Model}} & \multicolumn{3}{c}{Pneumonia Lin. Prob.} & \multicolumn{2}{c}{COVID} &\multicolumn{1}{c}{Pneumothorax} \\ \cmidrule(lr){2-4}\cmidrule(lr){5-6}\cmidrule(lr){7-7}
    & \multicolumn{1}{c}{Avg.} & \multicolumn{1}{c}{Single} & \multicolumn{1}{c}{Multiple} & \multicolumn{1}{c}{Lin. Prob.} & \multicolumn{1}{c}{Finetune} & \multicolumn{1}{c}{Finetune} \\ \midrule
    Random & \score{4.0}{2.7} & \score{7.5}{5.0} & \score{21.7}{14.5} & \score{7.1}{0.4} &\score{36.4}{2.2} & \score{31.8}{0.3} \\
    ImageNet~\cite{deng2009imagenet} & \score{43.5}{0.1} & \score{38.2}{0.1} & \score{58.1}{0.5} 
 & \score{24.7}{12.5} & \score{45.2}{0.2} & \score{42.8}{0.9} \\
    CLIP~\cite{2020_clip} & \score{52.0}{0.0} & \score{49.1}{0.1} & \score{\underline{66.9}}{0.1} & \score{48.2}{0.3} & \score{51.6}{1.7} & \score{43.2}{1.0} \\ 
    ConVIRT~\cite{zhang2022contrastive} & \score{52.2}{0.0} & \score{48.9}{0.1} & \score{66.5}{0.4} & \score{43.9}{0.0} & \score{49.0}{4.4} & \score{43.5}{0.9} \\
    LoVT~\cite{muller2021joint} & \score{\underline{52.4}}{0.0} & \score{\underline{49.1}}{0.2} & \score{66.9}{0.1} & \score{\underline{49.7}}{0.2} & \score{50.6}{5.0} & \score{44.3}{0.6} \\
    \textbf{\ours} & \score{\textbf{52.9}}{0.0} & \score{\textbf{49.3}}{0.4} & \score{\textbf{67.4}}{0.5} & \score{\textbf{51.5}}{0.3} & \score{\textbf{53.0}}{1.3} & \score{\underline{45.0}}{0.3} \\ %
    \midrule
    GLoRIA~\cite{Huang_2021_ICCV} \dag & \score{49.4}{0.0} & \score{46.9}{0.2} & \score{64.6}{0.1} & \score{35.6}{0.1} & \score{\underline{51.8}}{3.4} & \score{\textbf{45.1}}{0.3} \\ 
    BioViL-L~\cite{boecking2022making} \dag & \score{42.6}{0.0} & \score{40.0}{0.0} & \score{58.1}{0.3} & \score{34.2}{0.2} & \score{45.0}{2.0} & \score{39.2}{0.7} \\
    \bottomrule
  \end{tabular}

}
\end{table}

\parahead{Results}
As shown in \autoref{tab:main}, \OURS achieves the best segmentation performance on all downstream tasks, except for pneumothorax segmentation, where its performance is still comparable to that of the best-performing models. 
Notably, we observe a significant performance gain of \OURS on the COVID Rural dataset of limited size for both linear probing and fine-tuning tasks.
In practice, a limited data regime is common, which highlights the benefit of our approach.
Upon subgroup analysis by the number of ground truths (GTs) in the Pneumonia linear probing, we observe a greater enhancement over baseline with multiple GTs compared to one GT. 
This indicates our similarity-based approach outperforms distance-based local contrast learning~\cite{muller2021joint} in capturing dispersed GTs.
The table reveals that GLoRIA and BioVIL exhibit poorer performance than the other baselines, which could be attributed to their assessment of the downstream task at a resolution different from the pre-training.

\subsection{Phrase Grounding}
Next, we conduct phrase grounding experiments that evaluate the ability to comprehend the vision-language relationship.
Specifically, we evaluate how much the model accurately identifies the locations of clinical findings in an image that corresponds with a provided text query.

\parahead{Datasets}
The \textbf{MS-CXR Local Alignment} dataset~\cite{boecking2022making} is a subset of the MIMIC-CXR dataset~\cite{2019_mimic_cxr}, curated for phrase grounding by board-certified radiologists.
The dataset comprises 1,162 image-sentence pairs of bounding boxes for eight different types of lesions, along with their corresponding phrases. 
To perform phrase grounding, we utilize a phrase query from the dataset to identify which regions of the image have a local representation that closely matches the representation of the query text.

\parahead{Metric}
To quantitatively analyze local alignment, \cite{boecking2022making} proposes using the contrast-to-noise ratio (CNR) with the similarity between the local representation of a given text query and the visual features.
Specifically, CNR measures the discrepancy between the similarity of interior and exterior of the bounding box region and is computed by $|\mu_{\bs_{in}} - \mu_{\bs_{out}}| / \sqrt{(\sigma_{\bs_{in}}^{2} + \sigma_{\bs_{out}}^{2})}$, where $\bs=\text{sim}(\bz_l^\cT, \bz_l^\cI)$ and $\mu_X$ and $\sigma_X$ denote mean and variance, respectively.

However, the absolute value used in the original formulation of CNR can make it challenging to determine whether a high CNR value is due to high similarity inside or outside the bounding box. 
To address this issue, we also report a non-absolute version of CNR, i.e., $(\mu_{\bs_{in}} - \mu_{\bs_{out}}) / \sqrt{(\sigma_{\bs_{in}}^{2} + \sigma_{\bs_{out}}^{2})}$. 
This version provides a measure of whether the similarity inside the box is higher than outside the box. To the best of our knowledge, it has not yet been proposed in any literature.

\begin{table}[t]
\centering
\caption{
    \textbf{Evaluation with phrase grounding task.}
    We report non-absolute and absolute (shown in parenthesis) versions of CNR.
    ``Scratch'' is initialized with ImageNet and ClinicalBERT pre-trained weight without any joint training. 
    We also report three subgroups grouped by lesion due to space constraint.
    Please see the Supplementary Material for all subgroup results.
    \dag : We report the results from the previous study~\cite{boecking2022making}, which only includes absolute CNR. 
}
\label{tab:ms_cxr_lesion}
    \centering
    \begin{tabular}{lrrrr}
    \toprule
        \multicolumn{1}{c}{\multirow{2}[2]{*}{Model}} & \multicolumn{1}{c}{\multirow{2}[2]{*}{Avg.}} & \multicolumn{3}{c}{Findings} \\ \cmidrule(lr){3-5}
         & & \multicolumn{1}{c}{PE} & \multicolumn{1}{c}{PNA} & \multicolumn{1}{c}{PTX} \\ \midrule
        Scratch~\cite{alsentzer2019publicly,deng2009imagenet} & 0.118 (0.370) & 0.530 (0.591) & 0.187 (0.381) & -0.124 (0.425) \\ 
        CLIP~\cite{2020_clip} & -0.120 (0.340) & -0.311 (0.422) & -0.175 (0.328) & 0.072 (0.355) \\
        ConVIRT~\cite{zhang2022contrastive} & -0.015 (0.434) 
 & 0.375 (0.529) & -0.113  (0.427) & -0.085 (0.489) \\
        LoVT~\cite{muller2021joint} & 0.006 (0.366) & 0.190 (0.415) & 0.151 (0.356) & 0.030 (0.424) \\
        \ours & \textbf{1.117} (1.150) & \textbf{0.662} (0.677) & \textbf{1.478} (1.478) & \textbf{0.247} (0.396) \\  \midrule
        GLoRIA~\cite{Huang_2021_ICCV} \dag & - (0.930) & - (1.200) & - (1.180) & - (0.470) \\
        BioViL-L~\cite{boecking2022making} \dag & - (1.142) & - (1.500) & - (1.190) & - (0.740) \\  \bottomrule
    \end{tabular}

\end{table}

\begin{figure}[t]
\centering
\includegraphics[width=0.9\textwidth]{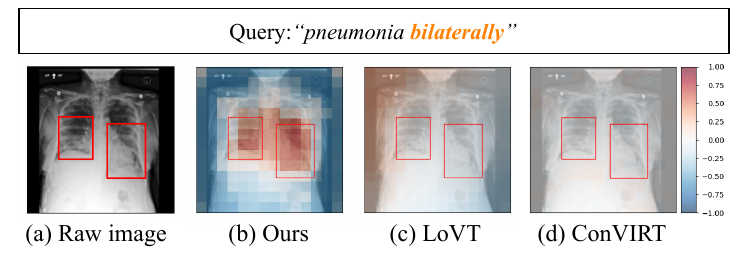}
\caption{
\textbf{Visualization of phrase grounding examples} For each given text queries, we visualize image and ROIs (with red boxes) related to the text on the left, and similarity map of image and text representations. We compare \OURS with two baselines: LoVT and ConVIRT. %
}\centering
\label{fig:cnr_example}
\end{figure}

\parahead{Result}
\autoref{tab:ms_cxr_lesion} demonstrates that \OURS exhibits the highest absolute and non-absolute CNR values, suggesting that preserving intra-modal local similarity can enhance the understanding of the vision-language correlation.
Note that the CNR is calculated on a per-pixel basis, but due to the different input sizes of GLORIA and BIOVIL-L and the resulting differences in feature map sizes, it is challenging to make a fair comparison.
When analyzing results by lesion, CNR values of \OURS are notably high for pneumonia which is typically located within the lung, and relatively low for pleural effusion and pneumothorax located mainly near the lung border.
Notably, all methods, except for \OURS, show a substantial disparity between their absolute and non-absolute CNR values, indicating that numerous samples exhibit high similarity between text and the exterior of the ground truth (GT). 
This is not ideal for localizing the actual ROI and suggests that evaluating local alignment with absolute CNR may be overstated.

As shown in \autoref{fig:cnr_example}, \OURS successfully encodes both text and text-related regions into similar embeddings.
Furthermore, our method effectively separates the embeddings of the foreground from those of the background.
In contrast, LoVT and ConVIRT fail to distinguish them and encode all regions, regardless of ROI, with near-zero similarities.
We claim that this separation helps segmentation by allowing the model to focus solely on the foreground.

\section{Conclusion}%
In this paper, we address the information loss of intra-modal locality during dimension reduction from unimodal space to cross-modal space. 
We have demonstrated that existing approaches to VLP in CXR do not consider locality sufficiently, resulting in poor downstream localization performance and misalignment in local embeddings of image and text. 
To overcome this challenge, we presented a novel approach called \OURS that employs a local contrastive objective to preserve intra-modal locality.
We train and evaluate the method on three datasets containing three different chest abnormalities. 
We demonstrate through our experiments that \OURS outperforms existing state-of-the-art VLP methods, which struggle to match local image patches with text phrases. By making VLP methods for CXR more location-aware, we take a step towards making practical CAD systems for CXR applications that are more accurate yet more efficient to train for various downstream localization tasks.

\parahead{Limitation and Discussion}
While we observe that our model encodes the text query with high similarity to the query-related regions, we also find that the model tends to broadly attend to the entire lung rather than focusing solely on the ROI for some cases of small lesions such as pneumothorax. 
(See Supplementary Material for more examples.) 
Nevertheless, our model achieves high phrase grounding performance, which leads to better localization performance in downstream tasks. 
Thus, we expect that the model which can similarly encode only exact locations to the text query would lead to better performance and reserve this as a future direction.

\bibliographystyle{splncs04}
\bibliography{refs}

\renewcommand{\theequation}{S\arabic{equation}}
\renewcommand{\thetable}{S\arabic{table}}
\renewcommand{\thefigure}{S\arabic{figure}}
\setcounter{equation}{0}
\setcounter{table}{0}
\setcounter{figure}{0}
\newpage
\begin{center}
    \Large\textbf{Supplementary Material}
\end{center}
\begin{table}[h]
\centering
\caption{
    \textbf{Evaluation with phrase grounding task.}
    We report non-absolute and absolute (shown in parenthesis) values of contrast-to-noise ratio (CNR).
    We conduct subgroup analysis for lesions and the number of ground-truth (GT) bounding boxes in the sample. \OURS outperforms the baselines in capturing regions of interest (ROIs) at multiple locations and shows consistently superior phrase grounding performance for five different clinical findings.
    \dag : We report the results from the previous study~\cite{boecking2022making}. %
    ATL: Atelectasis, CMG: Cardiomegaly, CONS: Consolidation, ED: Edema, LO: Lung Opacity, PE: Pleural effusion, PNA: Pneumonia, PTX: Pneumothorax.
}
    \centering
    \begin{tabular}{lrrrrrrrr}
    \toprule
        \multicolumn{1}{c}{\multirow{2}[2]{*}{Model}} & \multicolumn{1}{c}{\multirow{2}[2]{*}{Avg.}} & \multicolumn{5}{c}{Findings} & \multicolumn{2}{c}{\# GT boxes} \\ \cmidrule(lr){3-7}\cmidrule(lr){8-9}
         & & \multicolumn{1}{c}{ATL} & \multicolumn{1}{c}{CMG} & \multicolumn{1}{c}{CONS} & \multicolumn{1}{c}{ED} & \multicolumn{1}{c}{LO} & \multicolumn{1}{c}{Single} & \multicolumn{1}{c}{Multiple}  \\ \midrule
        Scratch & 0.118 & 0.164 & 0.141 & 0.070 & 0.023 & 0.198 & 0.096 & 0.191 \\ 
        ~ & (0.370) & (0.331) & (0.293) & (0.310) & (0.307) & (0.390) & (0.380) & (0.339) \\ 
        CLIP & -0.120 & -0.241 & -0.092 & -0.257 & -0.092 & -0.195 & -0.115 & -0.138 \\ 
        ~ & (0.340) & (0.410) & (0.271) & (0.403) & (0.307) & (0.378) & (0.352) & (0.299) \\ 
        ConVIRT & -0.015 & 0.026 & 0.094 & -0.163 & -0.446 & -0.061 & 0.006 & -0.082 \\ 
        ~ & (0.434) & (0.432) & (0.304) & (0.507) & (0.566) & (0.531) & (0.443) & (0.405) \\ 
        LoVT & 0.006 & 0.136 & -0.291 & 0.146 & 0.145 & 0.237 & -0.035 & 0.143 \\ 
        ~ & (0.366) & (0.326) & (0.328) & (0.366) & (0.241) & (0.416) & (0.396) & (0.271) \\ 
        Ours & \textbf{1.117} & \textbf{1.311} & \textbf{1.466} & \textbf{1.429} & \textbf{1.383} & \textbf{1.288} & \textbf{1.068} & \textbf{1.275} \\ 
        ~ & (1.150) & (1.314) & (1.466) & (1.429) & (1.385) & (1.288) & (1.109) & (1.281) \\  \midrule
        GLoRIA \dag & \multicolumn{1}{c}{-} & \multicolumn{1}{c}{-} & \multicolumn{1}{c}{-} & \multicolumn{1}{c}{-} & \multicolumn{1}{c}{-} & \multicolumn{1}{c}{-} & \multicolumn{1}{c}{-} & \multicolumn{1}{c}{-} \\ 
        ~ & (0.930) & (0.980) & (0.530) & (1.380) & (0.660) & (1.050)  & \multicolumn{1}{c}{(-)} &  \multicolumn{1}{c}{(-)}\\ 
        BioViL-L \dag & \multicolumn{1}{c}{-} & \multicolumn{1}{c}{-} & \multicolumn{1}{c}{-} & \multicolumn{1}{c}{-} & \multicolumn{1}{c}{-} & \multicolumn{1}{c}{-} & \multicolumn{1}{c}{-} & \multicolumn{1}{c}{-} \\
        ~ & (1.142) & (1.170) & (0.950) & (1.450) & (0.960) & (1.190) & \multicolumn{1}{c}{(-)} &  \multicolumn{1}{c}{(-)} \\  \bottomrule
    \end{tabular}
\end{table}

\begin{table}[h]
    \centering
    \caption{
    \textbf{Detailed hyperparameters for training.} %
    }\label{tab:hp_search}
    \begin{tabular}{lrrrrrr}
    \toprule
        \multicolumn{1}{c}{\multirow{2}[2]{*}{Hyperparameter}} & \multicolumn{1}{c}{\multirow{2}[2]{*}{\Pretrain ing}} & \multicolumn{1}{c}{Pneumonia} & \multicolumn{2}{c}{COVID} &\multicolumn{1}{c}{Pneumothorax} \\ \cmidrule(lr){3-3}\cmidrule(lr){4-5}\cmidrule(lr){6-6}
        & & \multicolumn{1}{c}{Lin. Prob.} & \multicolumn{1}{c}{Lin. Prob.} & \multicolumn{1}{c}{Finetune} & \multicolumn{1}{c}{Finetune} \\ \midrule
        Learning rate (LR) & 1.00E-04 & 1.00E-02 & 1.00E-02 & 2.00E-04 & 5.00E-05 \\
        LR scheduler & Cosine & Plateau & Plateau & Plateau & Step(30, 40) \\
        Epoch & 50 (w/ early stop) & 50 & 100 & 100 & 50 \\
        Batch size & $16\cdot8$(GPU) & 64 & 64 & 64 & 64 \\ \midrule
        Rotation angle & $(-20^\circ, 20^\circ)$ & $(-45^\circ, 45^\circ)$ & $(-10^\circ, 10^\circ)$ & $(-45^\circ, 45^\circ)$ & $(-45^\circ, 45^\circ)$ \\
        Scaling factor & (0.95, 1.05) & (0.9, 1.1) & (0.9, 1.1) & (0.9, 1.1) & (0.9, 1.1) \\
        Color jitter & (0.6, 1.4) & - & - & - & - \\
        Horizontal flip & 0.5 & - & - & - & - \\
        Random crop scale & (0.6, 1.0) & - & - & - & - \\
        Gaussian blur $\sigma$ & (0.1, 3.0) & - & - & - & - \\ \midrule
        Training time & 10 h & 2 h & 20 min & 50 min & 3 h \\ \bottomrule
    \end{tabular}
\end{table}

\begin{figure}[t]
\caption{
\textbf{Visualization of extensive phrase grounding examples using MS-CXR dataset~\cite{boecking2022making}.} 
\OURS tends to capture similarity between broad regions in both lungs, while two baselines highlight regions outside ROIs.
}\centering
\label{fig:cnr_example_supp}
\centering
\includegraphics[width=0.75\textwidth]{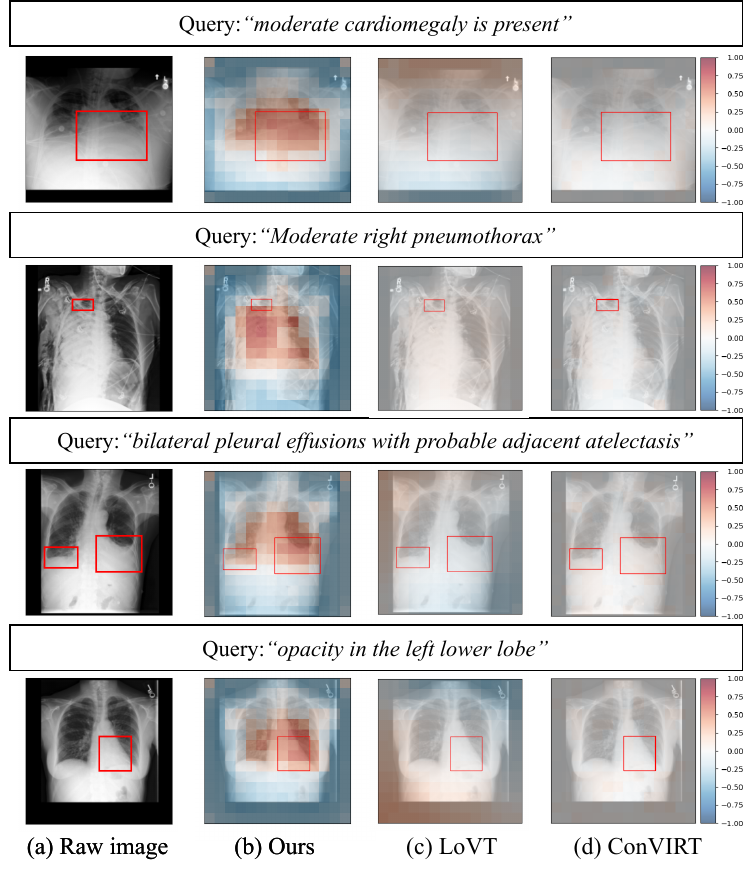}
\end{figure}

\end{document}